\documentclass{article}

\usepackage{tabularx}
\usepackage{amssymb}

\usepackage{arxiv}

\usepackage[utf8]{inputenc} 
\usepackage[T1]{fontenc}    
\usepackage{hyperref}       
\usepackage{url}            
\usepackage{booktabs}       
\usepackage{amsfonts}       
\usepackage{nicefrac}       
\usepackage{microtype}      
\usepackage{graphicx}
\graphicspath{ {./images/} }

\title{HLER: Human-in-the-Loop Economic Research via Multi-Agent Pipelines for Empirical Discovery}

\author{
 Chen Zhu \\
  China Agricultural University \\
  \texttt{zhuchen@cau.edu.cn} \\
   \And
 Xiaolu Wang \\
  China Agricultural University \\
  \texttt{wangxiaolu77@cau.edu.cn} \\
}

\begin{document}
\maketitle
\begin{abstract}
Large language models (LLMs) have enabled agent-based systems that attempt to automate scientific research workflows. Existing approaches often pursue fully autonomous discovery, in which AI systems generate research ideas, conduct experiments, and produce manuscripts with minimal human involvement. However, empirical research in economics and the social sciences poses additional challenges: research questions must be grounded in available datasets, identification strategies require careful design, and human judgment remains essential for evaluating economic significance. We introduce HLER (Human-in-the-Loop Economic Research), a multi-agent architecture that supports empirical research automation while preserving critical human oversight. The system orchestrates specialized agents for data auditing, data profiling, hypothesis generation, econometric analysis, manuscript drafting, and automated review. A central design principle is \textit{dataset-aware hypothesis generation}, in which candidate research questions are constrained by dataset structure, variable availability, and distributional diagnostics. This prevents infeasible or hallucinated hypotheses that frequently arise in unconstrained LLM ideation. HLER further introduces a two-loop research architecture: a \textit{question quality loop} in which candidate hypotheses are generated, screened for feasibility, and selected by a human researcher; and a \textit{research revision loop} in which automated review triggers re-analysis and manuscript revision. Human decision gates are embedded at key stages, including question selection and publication approval, allowing researchers to steer the automated pipeline. We evaluate the framework on three empirical datasets, including longitudinal survey data. Across 14 pipeline runs, dataset-aware hypothesis generation produces feasible research questions in 87\% of cases, compared with 41\% under unconstrained generation, and the system successfully completes end-to-end empirical manuscripts at an average API cost of \$0.8-\$1.5 per run. A detailed case study on the China Health and Nutrition Survey illustrates the full workflow from data inspection to a revised draft manuscript. These results suggest that Human-AI collaborative pipelines may provide a practical path toward scalable empirical research. The HLER working paper repository is publicly available at \url{https://github.com/maxwell2732/hler-working-papers}.
\end{abstract}

\section{Introduction}

Recent advances in large language models (LLMs) have substantially improved automated scientific writing, including summarization, drafting, and the generation of coherent technical narratives \cite{achiam2023gpt, anthropic2024model, taylor2022galactica}. Frontier model reports document broad improvements in academic and professional task performance \cite{achiam2023gpt, anthropic2024model}, and domain-targeted scientific LLMs illustrate the feasibility of using language models as interfaces for scientific knowledge \cite{taylor2022galactica}. At the same time, most existing LLM-based workflows remain primarily text-centric: they can draft plausible prose but struggle to reliably perform the end-to-end steps required for credible empirical research, particularly in data-driven social sciences.

One widely documented limitation is hallucination, generation models can produce fluent but unsupported claims, which motivates verification, constraint, and critique mechanisms when models are used in research settings \cite{ji2023survey}. A second limitation is that empirical research requires interaction with external artifacts (datasets, code, econometric software, and intermediate results). Agentic paradigms such as ReAct demonstrate that reliable task completion often requires coupling LLM reasoning with tool-mediated actions that query external sources and reduce error propagation \cite{yao2022react}. These limitations imply that simply generating a paper-like narrative is insufficient for empirical disciplines where credibility depends on data validation, transparent estimation, and robustness.

A growing body of work explores autonomous research, vibe research, and ``AI scientist'' pipelines that extend LLMs beyond writing to include idea generation, code execution, and automated review, largely in machine learning research environments \cite{lu2024ai}. These systems provide an important precedent for end-to-end automation, but they also highlight a gap for empirical economics and related social sciences: credible empirical contribution depends on dataset feasibility checks, econometric identification choices, and reproducible workflows, not only on experimental iteration in simulated or ML-centric settings.

Empirical economics is inherently procedural and iterative. A typical study requires (i)~validating dataset structure and variable definitions, (ii)~selecting feasible questions that match observed variation, (iii)~specifying an identification strategy, (iv)~estimating models, and (v)~subjecting results to critical scrutiny prior to dissemination. Evidence from replication audits suggests that even when papers are required to provide data and code, reproducing published economics findings often fails or requires substantial author assistance, underscoring the fragility of empirical workflows \cite{chang2018economics}. More broadly, social-scientific conclusions can be highly sensitive to analytic choices: even holding data and the research question fixed, different teams can arrive at meaningfully different estimates due to modeling decisions \cite{silberzahn2018many}.

This paper introduces the Human-in-the-Loop Economic Research (HLER) pipeline, a multi-agent architecture designed to automate the empirical research workflow while preserving key human oversight. The system orchestrates specialized agents that sequentially perform data auditing, data profiling, research question generation, econometric analysis, manuscript drafting, and automated review. Our orchestration approach is inspired by recent multi-agent frameworks that coordinate specialized LLM roles and allow flexible mixtures of tool use and human inputs \cite{wu2024autogen}. To improve feasibility and reduce hallucinated hypotheses, HLER implements \textit{dataset-aware question generation}, constraining candidate questions by dataset structure, variable availability, missingness, and distributional diagnostics rather than unconstrained ideation.

A central design feature of HLER is a two-loop research architecture. The \textit{question quality loop} generates candidate hypotheses, screens them for feasibility using dataset diagnostics, and presents them to a human researcher for selection. The \textit{research revision loop} subjects the resulting manuscript to automated review; the reviewer agent may request additional robustness checks or improved exposition, triggering re-analysis and manuscript revision. This iterative design is motivated by evidence that structured self-feedback can substantially improve model outputs without additional training \cite{madaan2023self}. By integrating data constraints, econometric execution, and iterative critique within a single orchestrated pipeline, HLER aims to extend LLM utility from automated writing toward scalable, human-supervised empirical discovery in economics.

Our main contributions are:
\begin{enumerate}
    \item A \textit{dataset-aware hypothesis generation} mechanism that conditions research question generation on dataset audits and statistical profiles, reducing infeasible hypotheses from 59\% to 13\% compared with unconstrained LLM ideation.
    \item A \textit{two-loop research architecture} consisting of a question quality loop (generation $\rightarrow$ screening $\rightarrow$ human selection) and a research revision loop (review $\rightarrow$ re-analysis $\rightarrow$ revision), with empirical evidence that reviewer scores improve across revision iterations.
    \item An end-to-end evaluation across three empirical datasets and 14 pipeline runs, demonstrating that the system can reliably produce complete research manuscripts at an average cost of \$0.8-\$1.5 per run.
\end{enumerate}

\section{Related Work}
\label{sec:related_work}

\subsection{LLM Agents for Scientific Discovery}

Recent advances in large language models have led to growing interest in using AI systems as autonomous or semi-autonomous researchers. Several studies propose multi-agent architectures capable of generating hypotheses, implementing experiments, and producing scientific manuscripts. The AI Scientist introduces an end-to-end framework in which LLM agents autonomously generate research ideas, write code, analyze results, and produce full scientific papers, followed by an automated review process \cite{lu2024ai}. More broadly, a growing literature explores agent-based research automation, where multiple specialized agents coordinate to perform different components of the scientific workflow \cite{liu2025vision}.

Related work has also explored domain-specific autonomous research pipelines. BioResearcher proposes a modular multi-agent system for biomedical research workflows, including literature processing, experiment design, and protocol generation \cite{luo2025intention}. AutoLabs attempts to automate experimental workflows in chemistry using self-correcting agents and structured reasoning \cite{panapitiya2025autolabs}. While these systems demonstrate the feasibility of AI-assisted research, most existing architectures focus on computational or experimental sciences, where experiments can be directly executed in simulation or laboratory environments.

\subsection{Autonomous Research Systems in Economics}

In economics and policy evaluation, research automation poses different challenges. Empirical social science requires not only hypothesis generation and analysis but also careful handling of observational datasets, identification strategies, and causal inference.

A prominent recent effort is the Autonomous Policy Evaluation (APE) project developed by the Social Catalyst Lab at the University of Zurich \cite{ape_github_2026}. APE explores whether an AI system can autonomously generate empirical economics research at scale using publicly available data. The system produces full research papers end-to-end by identifying research questions, retrieving public datasets, writing econometric analysis code (primarily in R), generating figures and tables, and drafting \LaTeX\ manuscripts. APE papers are evaluated through a tournament framework in which AI-generated papers are compared against human-written research using automated LLM referees.

However, the APE framework intentionally emphasizes full automation, where most analysis code and manuscripts are generated without direct human intervention. While this approach enables large-scale experimentation, it also highlights challenges related to hallucinated hypotheses and limited feasibility checks.

\subsection{Human-in-the-Loop Research Automation}

An alternative approach emphasizes Human-AI collaboration rather than full automation \cite{dautenhahn1998art, takerngsaksiri2025human}. In complex empirical fields such as economics, research design often requires domain expertise, judgment about identification strategies, and evaluation of economic significance.

HLER builds on this perspective by combining autonomous agent execution with explicit human decision gates. The framework differs from prior systems in three key ways. First, HLER explicitly integrates human decision gates within the research workflow, while systems such as APE primarily explore the feasibility of fully autonomous pipelines. Second, HLER introduces a data-driven question generation stage based on dataset auditing and profiling, which constrains hypothesis generation using information about variable availability, missingness, and distributional properties. Third, HLER incorporates a structured two-loop research architecture: a question quality loop (generation $\rightarrow$ feasibility screening $\rightarrow$ human selection) and a research revision loop (automated review $\rightarrow$ re-analysis $\rightarrow$ manuscript revision), designed to approximate the iterative review processes typical of empirical research.

Table~\ref{tab:system_comparison} summarizes the key differences between HLER and existing AI research automation systems.

\begin{table}[t]
\centering
\footnotesize
\caption{Comparison of HLER with existing AI research automation systems.
$\checkmark$ = supported, $\circ$ = partially supported, $-$ = not supported. ``LLM Agents'' refers to general-purpose multi-agent frameworks such as AutoGen \cite{wu2024autogen} and similar systems \cite{liu2025vision}. ``Writing Tools'' refers to LLM-based writing assistants without empirical analysis capabilities.}
\label{tab:system_comparison}

\begin{tabular}{lccccc}
\toprule
\textbf{Feature} 
& \textbf{HLER} 
& \textbf{AI Scientist} 
& \textbf{APE} 
& \textbf{LLM Agents} 
& \textbf{Writing Tools} \\
\midrule

Domain
& Empirical Econ
& ML
& Policy
& General
& Writing \\

Hypothesis generation
& $\checkmark$
& $\checkmark$
& $\checkmark$
& $\circ$
& - \\

Empirical / experimental analysis
& $\checkmark$
& $\checkmark$
& $\checkmark$
& -
& - \\

Automated paper drafting
& $\checkmark$
& $\checkmark$
& $\checkmark$
& -
& $\checkmark$ \\

Automated review / critique
& $\checkmark$
& $\checkmark$
& $\circ$
& -
& - \\

Human-in-the-loop
& $\checkmark$
& -
& $\circ$
& $\circ$
& $\checkmark$ \\

\midrule
\multicolumn{6}{l}{\textbf{HLER-specific capabilities}} \\

Data-aware hypothesis generation
& $\checkmark$
& -
& -
& -
& - \\

Data auditing / profiling
& $\checkmark$
& -
& -
& -
& - \\

\bottomrule
\end{tabular}

\end{table}

\section{System Design and Implementation}
\label{sec:system}

HLER (Human-in-the-Loop Economic Research) is a multi-agent pipeline that decomposes empirical research into structured stages executed by specialized agents coordinated by a central orchestrator. This section describes both the conceptual architecture and its implementation. Multi-agent architectures have recently been used to automate aspects of scientific discovery, including hypothesis generation, experiment design, and iterative refinement \cite{ai_scientist_v2_2025, robin_2025}. HLER adapts this paradigm to the structure of economic empirical research.

\subsection{Architecture Overview}

\begin{figure}[t]
  \centering
  \includegraphics[width=0.85\linewidth]{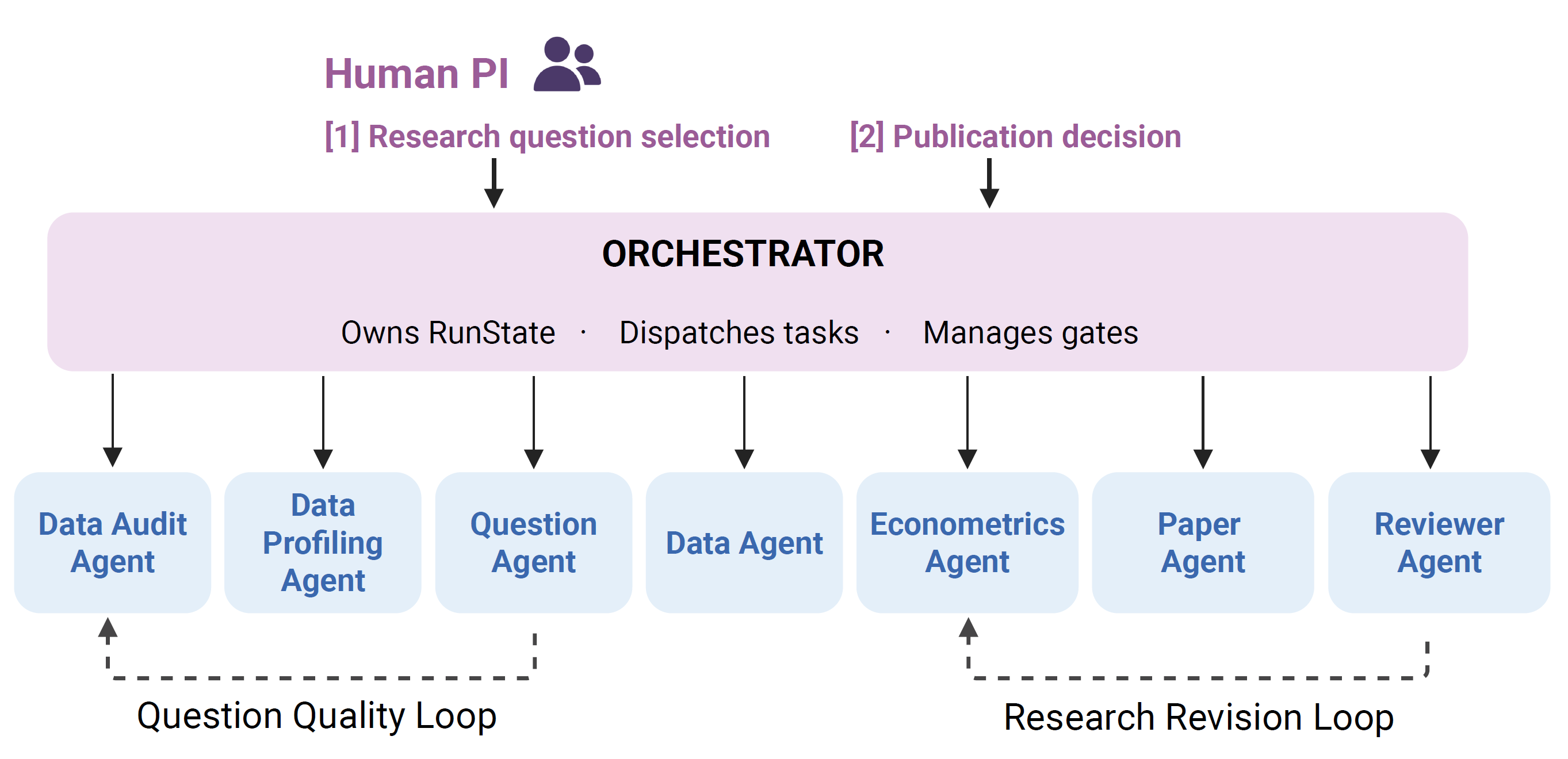}
  \caption{Architecture of the HLER system. The orchestrator coordinates a multi-agent empirical research pipeline. The core workflow includes data auditing, data profiling, question generation and screening, econometric analysis, manuscript drafting, and automated critique. Two feedback loops are shown: the \textit{question quality loop} (left), in which candidates are generated and screened before human selection; and the \textit{research revision loop} (right), in which the reviewer agent triggers re-analysis and manuscript revision through the Econometrics and Paper agents. Human decision gates allow researchers to select research questions and approve final outputs.} 
  \label{fig:hler_architecture}
\end{figure}

Figure~\ref{fig:hler_architecture} illustrates the overall architecture. At the top level, a human principal investigator (PI) interacts with the system through two decision gates: research question selection and publication approval. Between these gates, a central orchestrator coordinates the activities of multiple specialized agents.

The orchestrator maintains a shared state object (\texttt{RunState}) that records intermediate outputs from each stage of the pipeline. Tasks are dispatched sequentially to agents according to the predefined workflow:

\begin{center}
\textsc{Data Audit} $\rightarrow$ \textsc{Data Profiling} $\rightarrow$ \textsc{Questioning} $\rightarrow$ 
\textsc{Data Collection} $\rightarrow$ \textsc{Analysis} $\rightarrow$ \textsc{Writing} $\rightarrow$ 
\textsc{Self-Critique} $\rightarrow$ \textsc{Review}
\end{center}

Each agent produces structured outputs that are stored in the shared state and passed to the next stage. The final stage returns either a completed research draft or revision requests, which may trigger an additional pipeline iteration.

\subsection{Agent Roles}

HLER consists of seven specialized agents, each responsible for a distinct component of the research workflow.

\paragraph{Data Audit Agent.}
The \texttt{DataAuditAgent} validates the structure of available datasets before any research questions are generated. The agent reads dataset headers and schema information to construct an inventory of available variables and data sources. This step prevents the system from proposing hypotheses that require variables absent from the dataset.

\paragraph{Data Profiling Agent.}
The \texttt{DataProfilingAgent} analyzes the statistical properties of the dataset, including summary statistics, missingness patterns, variable distributions, and pairwise correlations. The agent also identifies potential endogeneity risks and suggests variable transformations. These diagnostics are summarized in a structured \texttt{DataProfile} object that informs subsequent hypothesis generation.

\paragraph{Question Agent.}
The \texttt{QuestionAgent} generates candidate empirical research questions using a large language model constrained by the dataset audit and data profile. Unlike conventional LLM brainstorming, the hypothesis generation process is explicitly conditioned on dataset availability and statistical properties, reducing the frequency of infeasible research questions.

\paragraph{Data Agents.}
Data collection agents retrieve relevant data from public APIs (e.g., World Bank, FRED) and local datasets. The retrieved data are merged and prepared for analysis.

\paragraph{Econometrics Agent.}
The \texttt{EconometricsAgent} constructs an empirical analysis plan and executes the appropriate econometric models. Depending on the dataset structure, the agent may estimate ordinary least squares (OLS), fixed-effects panel models, difference-in-differences designs, or event-study specifications. Results are exported as tables, figures, and structured output summaries.

\paragraph{Paper Agent.}
The \texttt{PaperAgent} generates a full research manuscript based on the econometric results. The draft includes an abstract, introduction, methodology, results, and discussion sections, following standard academic paper conventions. Drafts are versioned (e.g., \texttt{draft\_v1.md}, \texttt{draft\_v2.md}) to support iterative revision.

\paragraph{Reviewer Agent.}
The \texttt{ReviewerAgent} evaluates the draft across several criteria including novelty, identification credibility, data quality, clarity, and policy relevance. The agent produces structured review reports with numerical scores (1-10 scale) and identifies specific areas requiring revision.

\subsection{Control Loops}
\label{sec:control_loops}

HLER incorporates two feedback loops that improve research quality.

\paragraph{Question Quality Loop.}
During hypothesis generation, the system produces multiple candidate research questions. These questions are evaluated using feasibility checks and preliminary statistical diagnostics. The human PI then selects the most promising research direction through the question-selection gate. If none of the candidates are satisfactory, the researcher can request a new generation round with modified constraints, making this a genuine iterative loop.

\paragraph{Research Revision Loop.}
After the manuscript is generated, the \texttt{ReviewerAgent} evaluates the paper and may request revisions. Each iteration of this loop consists of two phases: a \textit{production phase}, in which the \texttt{EconometricsAgent} performs additional analyses requested by the reviewer (e.g., robustness checks, alternative specifications) and the \texttt{PaperAgent} updates the manuscript; and a \textit{critique phase}, in which the \texttt{ReviewerAgent} re-evaluates the revised draft and determines whether further revisions are needed. This loop can repeat multiple times. In practice, we observe convergence within 2-4 iterations (see Section~\ref{sec:revision_dynamics}).

\subsection{Human Oversight}

Despite its high level of automation, HLER is designed as a human-in-the-loop system. Humans intervene at two key decision points:

\begin{enumerate}
\item \textbf{Research Question Selection} --- the human researcher selects the most promising hypothesis among the generated candidates, applying domain knowledge and research judgment.

\item \textbf{Publication Decision} --- the human decides whether the resulting paper is sufficiently credible and interesting to be finalized, acting as a final quality gate.
\end{enumerate}

This design balances automation with human judgment, ensuring that critical scientific decisions remain under researcher control while routine tasks such as data processing and manuscript drafting are automated.

\subsection{Implementation Details}

HLER is implemented in Python. All LLM reasoning is performed via the Anthropic API using Claude Sonnet 4.6, though the architecture is model-agnostic. Each agent is a Python class derived from a shared \texttt{BaseAgent} interface that standardizes task reception, shared context access, and result return. Agent outputs are represented using structured data objects (e.g., \texttt{ResearchQuestion}, \texttt{DataProfile}, \texttt{AnalysisResult}) defined in a shared message schema, enabling logging, validation, and reuse across pipeline stages.

The system supports both local datasets and external data sources. Local datasets are accessed through dataset-specific client modules that standardize file loading, metadata extraction, and variable inspection. The current prototype supports the China Health and Nutrition Survey (CHNS), the China Multi-Generational Panel Dataset (CMGPD-Liaoning), and the UK Biobank. Public macroeconomic data can be retrieved through external APIs such as the World Bank, FRED, and OpenAlex.

The econometric analysis stage combines LLM-generated analysis plans with programmatic statistical estimation using standard Python libraries (e.g., \texttt{statsmodels}, \texttt{linearmodels}). The current implementation supports OLS, fixed-effects panel regression, difference-in-differences, and event-study specifications. Research manuscripts are generated in Markdown format and compiled into PDF using Pandoc and \LaTeX.

\section{Experimental Evaluation}
\label{sec:evaluation}

We evaluate HLER along four dimensions: (1)~the feasibility of generated research questions under dataset-aware versus unconstrained generation, (2)~the end-to-end pipeline completion rate, (3)~the dynamics of the revision loop, and (4)~the computational cost.

\subsection{Experimental Setup}

All experiments use the implementation described in Section~\ref{sec:system} with Claude Sonnet 4.6 through the Anthropic API. We evaluate on three empirical datasets commonly used in economic research:

\begin{itemize}
\item China Health and Nutrition Survey (CHNS) --- longitudinal survey, 285 variables, 57{,}203 observations, 1989-2011
\item China Multi-Generational Panel Dataset (CMGPD-Liaoning) --- historical demographic records
\item UK Biobank --- large-scale biomedical dataset with socioeconomic variables
\end{itemize}

For each dataset, we executed the full HLER pipeline multiple times with different research domain constraints (e.g., labor economics, health economics, education). A total of 14 complete pipeline runs were conducted across the three datasets. The human researcher selected one research question per run from the generated candidates.

\subsection{Research Question Feasibility}
\label{sec:feasibility}

A central goal of HLER is to produce empirically feasible research questions. To evaluate this, we compare questions generated under two conditions:

\begin{itemize}
\item \textbf{Dataset-aware generation} (HLER default): The \texttt{QuestionAgent} receives the full dataset audit and statistical profile before generating hypotheses.
\item \textbf{Unconstrained generation} (ablation): The same LLM generates research questions given only a dataset name and broad domain description, without audit or profiling information.
\end{itemize}

Each generated question was evaluated against three feasibility criteria: (1)~the required variables exist in the dataset, (2)~the proposed empirical design is compatible with the dataset structure, and (3)~the question can be addressed using the supported econometric methods. Two researchers independently assessed each question; disagreements were resolved through discussion.

Table~\ref{tab:feasibility} reports the results. Across 14 runs, the dataset-aware mechanism generated 79 candidate questions, of which 69 (87\%) satisfied all three feasibility criteria. The unconstrained ablation generated 82 questions, of which only 34 (41\%) were fully feasible. The most common failure mode under unconstrained generation was proposing variables not present in the dataset (42\% of failures), followed by designs incompatible with the data structure (35\%).

\begin{table}[t]
\centering
\footnotesize
\caption{Feasibility of generated research questions under dataset-aware versus unconstrained generation across 14 pipeline runs on three datasets.}
\label{tab:feasibility}
\begin{tabular}{lcccc}
\toprule
\textbf{Condition} & \textbf{Questions} & \textbf{Feasible (\%)} & \textbf{Var.\ missing} & \textbf{Design incompat.} \\
\midrule
Dataset-aware (HLER) & 79 & 69 (87\%) & 4 & 6 \\
Unconstrained (ablation) & 82 & 34 (41\%) & 20 & 17 \\
\bottomrule
\end{tabular}
\end{table}

\subsection{End-to-End Pipeline Completion}

Of the 14 pipeline runs, 12 (86\%) completed the full workflow from dataset auditing through manuscript generation and review without manual intervention beyond the two planned decision gates. The two incomplete runs failed during the econometric analysis stage due to convergence issues in fixed-effects estimation on sparse subsamples. In both cases, the \texttt{EconometricsAgent} logged the error and the pipeline halted gracefully, allowing the researcher to adjust sample restrictions and re-run.

For completed runs, the system produced regression tables, figures, and structured summaries of empirical results. The \texttt{PaperAgent} generated full research manuscripts including abstract, introduction, empirical strategy, results, and discussion sections.

\subsection{Revision Loop Dynamics}
\label{sec:revision_dynamics}

To evaluate whether the research revision loop improves manuscript quality, we tracked the \texttt{ReviewerAgent}'s numerical scores across revision iterations for the 12 completed runs. The reviewer evaluates each draft on five dimensions (novelty, identification credibility, data quality, clarity, and policy relevance) using a 1-10 scale.

Table~\ref{tab:revision} reports the mean reviewer scores across iterations. Initial drafts received a mean overall score of 4.8 (SD = 0.9). After the first revision, scores improved to 5.9 (SD = 0.7). Most runs converged after 2-3 iterations, with final drafts scoring 6.3 (SD = 0.6) on average. The largest improvements were observed in clarity (+2.1 points on average) and identification credibility (+1.4), while novelty scores showed minimal change across iterations (+0.3), consistent with the expectation that the revision loop improves exposition and robustness rather than the underlying research contribution.

\begin{table}[t]
\centering
\footnotesize
\caption{Mean reviewer scores (1-10 scale) across revision iterations for 12 completed pipeline runs. SD in parentheses.}
\label{tab:revision}
\begin{tabular}{lccc}
\toprule
\textbf{Dimension} & \textbf{Draft v1} & \textbf{Draft v2} & \textbf{Final draft} \\
\midrule
Novelty & 5.2 (1.1) & 5.3 (1.0) & 5.5 (0.9) \\
Identification & 4.1 (1.2) & 5.2 (0.9) & 5.5 (0.8) \\
Data quality & 5.4 (0.8) & 5.8 (0.7) & 6.1 (0.6) \\
Clarity & 4.3 (1.0) & 5.8 (0.8) & 6.4 (0.7) \\
Policy relevance & 5.1 (0.9) & 5.5 (0.8) & 5.8 (0.7) \\
\midrule
\textbf{Overall} & 4.8 (0.9) & 5.9 (0.7) & 6.3 (0.6) \\
\bottomrule
\end{tabular}
\end{table}

The most frequent reviewer requests in the first iteration were: additional robustness checks (10 of 12 runs), clearer identification strategy discussion (8 of 12), and improved variable descriptions (7 of 12).

\subsection{Runtime and Cost}

A complete HLER pipeline run typically requires approximately 20-25 minutes on a standard research workstation. The average LLM API cost per run is \$0.8-\$1.5, depending on the number of generated research questions and revision iterations. This is substantially lower than the AI Scientist framework, which reports approximately \$6-\$15 per generated research paper \cite{lu2024ai}, primarily because HLER limits the scope of LLM calls to reasoning and text generation while delegating statistical computation to programmatic libraries.

\subsection{Case Study: Rural Women's Labor Outcomes in China}
\label{sec:case_study}

To illustrate the full HLER workflow, we present a detailed case study from a pipeline run on the China Health and Nutrition Survey (CHNS).

\paragraph{Setup.}
The researcher specified a research domain (labor economics) and a broad constraint focusing on rural women in China. The system then executed the full pipeline.

\paragraph{Dataset inspection.}
The \texttt{DataAuditAgent} identified the CHNS panel as the primary dataset, detecting 285 variables covering 1989-2011. The \texttt{DataProfilingAgent} computed summary statistics for 57{,}203 observations, identified variables with high missingness (e.g., income variables missing for 23\% of observations), and flagged potential endogeneity concerns between education and occupation variables.

\paragraph{Hypothesis generation and selection.}
The \texttt{QuestionAgent} generated eight candidate research questions. The \texttt{QuestionScreeningAgent} scored each candidate on feasibility and empirical tractability. Of the eight candidates, seven satisfied all feasibility criteria. The top-ranked question selected by the researcher was:

\begin{quote}
\textit{Does higher education attainment reduce the occupational gender gap among rural women in China?}
\end{quote}

\paragraph{Empirical analysis.}
The \texttt{EconometricsAgent} constructed a panel dataset with individual and wave identifiers and implemented a fixed-effects regression model. After applying sample restrictions, the analytic dataset contained 19{,}466 observations. The main specification estimated the relationship between education attainment and occupational category using within-individual variation.

\paragraph{Revision dynamics.}
The \texttt{ReviewerAgent} evaluated the initial draft (5{,}563 words) and identified concerns about reverse causality and identification credibility. The reviewer recommended event-study specifications, sensitivity analyses, and a clearer discussion of the fixed-effects identification assumption. Over three revision iterations, the \texttt{EconometricsAgent} performed additional analyses and the \texttt{PaperAgent} incorporated the new results and improved the exposition. The final manuscript contained 7{,}282 words. Reviewer scores improved from 4.6 (initial) to 6.5 (final), with the largest gains in identification credibility (3.2 $\rightarrow$ 5.8) and clarity (4.1 $\rightarrow$ 6.9).

\paragraph{Outcome.}
The final run produced a complete research manuscript together with replication files, regression outputs, and figures, demonstrating that HLER can execute the full empirical workflow while maintaining human oversight at key decision points.

\section{Discussion}

\subsection{Human-AI Co-Production of Empirical Research}

The results suggest a paradigm for empirical research in which AI agents automate routine tasks while human researchers retain control over key scientific decisions. Recent work on AI-driven scientific discovery has demonstrated the potential of multi-agent systems to automate hypothesis generation, experiment planning, and manuscript writing \cite{lu2024ai, robin_2025}. However, scientific research is fundamentally an interpretive activity that involves judgment, domain expertise, and value-based decisions.

Human-in-the-loop approaches offer an alternative design philosophy in which human expertise complements machine intelligence. Prior studies have shown that Human-AI collaboration can improve decision accuracy, reliability, and trust compared with fully automated approaches \cite{mosqueira2023hitl}. HLER operationalizes this paradigm through explicit human decision gates embedded within the research pipeline, allowing researchers to select promising hypotheses, evaluate intermediate results, and determine whether generated manuscripts should be pursued further.

\subsection{Scaling Empirical Research}

A key motivation for the HLER framework is the potential to scale empirical research. Many stages of empirical work, such as dataset inspection, variable screening, robustness checks, and manuscript drafting, are repetitive and time-consuming. By automating these steps, HLER allows researchers to explore a much larger space of potential research questions. This capability may be particularly valuable for data-rich fields such as health economics, labor economics, and development economics, where large observational datasets contain thousands of variables and complex panel structures. Multi-agent scientific systems have already demonstrated the ability to accelerate discovery in other domains \cite{ai_scientist_v2_2025}; our results suggest similar approaches can be adapted to empirical social science workflows.

\subsection{Researcher Degrees of Freedom and Ethical Considerations}
\label{sec:ethics}

The ability to rapidly generate and test many hypotheses raises important methodological concerns. When a system can produce dozens of candidate research questions for a single dataset, the risk of selective reporting and multiple-testing problems increases substantially. A researcher who evaluates many HLER-generated hypotheses but reports only the one with statistically significant results would be engaging in a form of automated p-hacking.

Several design features of HLER partially mitigate this concern. The question-selection gate creates a record of all generated hypotheses, not only those ultimately pursued. The structured pipeline logs all intermediate outputs, including generated questions that were not selected and analysis results from earlier iterations. These logs provide an audit trail that could support pre-registration-like transparency.

However, we emphasize that HLER does not solve the multiple-testing problem. Responsible use of the system requires that researchers apply the same standards of statistical discipline that govern conventional empirical work, including appropriate multiple-testing corrections when exploring many hypotheses and transparent reporting of the full set of questions considered. More broadly, the prospect of low-cost automated empirical research raises questions about the potential for flooding preprint servers and journals with large volumes of machine-generated manuscripts. We believe that human decision gates, particularly the publication approval stage, are a necessary safeguard, but the research community should develop norms and standards for AI-assisted empirical research as these tools become more widespread.

\subsection{Limitations}

The HLER system has several limitations. First, the quality of generated research questions depends on the available datasets and the variable inventory. While dataset-aware generation reduces infeasible questions, it does not guarantee that generated hypotheses are theoretically meaningful or novel. Second, the econometric analysis stage currently supports only a limited set of empirical designs. Many economic studies require instrumental variables, regression discontinuity designs, or structural models; expanding the supported methods is an important direction for future work. Third, the \texttt{ReviewerAgent} scores are generated by the same underlying LLM that produces the manuscripts, creating a potential circularity in the evaluation. While the reviewer and paper agents use different prompts and roles, an independent evaluation by human reviewers or a different LLM would provide stronger evidence of manuscript quality. Fourth, our evaluation is based on 14 pipeline runs across three datasets. While the results are consistent, larger-scale benchmarking across more datasets and research domains would strengthen the generalizability of our findings.

\subsection{Implications for Scientific Workflows}

The emergence of LLMs and multi-agent architectures may fundamentally change how scientific research is conducted. Future research workflows may involve collaborative systems in which AI agents assist researchers at multiple stages of the scientific process \cite{jr_ai_scientist_2025}. In this context, systems such as HLER should not be viewed as replacements for researchers but as tools that extend the capabilities of scientists, allowing them to explore larger hypothesis spaces and conduct more systematic robustness checks.

\section{Conclusion}

This paper introduces HLER (Human-in-the-Loop Economic Research), a multi-agent pipeline that automates the empirical research workflow while preserving human oversight in key scientific decisions. The system decomposes economic research into structured stages including dataset auditing, data profiling, hypothesis generation, econometric analysis, manuscript drafting, and automated review.

Our evaluation across three empirical datasets and 14 pipeline runs demonstrates that dataset-aware hypothesis generation produces feasible research questions in 87\% of cases, compared with 41\% under unconstrained LLM ideation. The system completes end-to-end research pipelines in 86\% of runs, and the research revision loop improves reviewer scores from 4.8 to 6.3 on average across 2-3 revision iterations. A detailed case study on labor outcomes for rural women in China illustrates the full workflow from data inspection to a revised draft manuscript.

Unlike fully autonomous research systems, HLER adopts an explicitly human-in-the-loop architecture. Human researchers intervene at key decision gates, ensuring that domain expertise and scientific judgment remain central to the research process \cite{mosqueira2023hitl}. This positions HLER as a collaborative tool rather than a replacement for researchers.

Future work could extend the HLER framework in several directions: expanding the econometric toolkit to include instrumental variables, regression discontinuity designs, and structural models; conducting large-scale benchmarking across diverse datasets and domains; incorporating independent human evaluation of generated manuscripts; and developing standards for transparent reporting in AI-assisted empirical research.

\bibliographystyle{unsrt}  
\bibliography{references}

\end{document}